%% file: main.tex
\documentclass[acmtog]{acmart} 

\acmSubmissionID{2145}

\copyrightyear{2026}
\acmYear{2026}
\setcopyright{cc}
\setcctype{by}
\acmConference[SA Conference Papers '26]{SIGGRAPH Asia 2026 Conference Papers}{December 01--04, 2026}{Kuala Lumpur, Malaysia}
\acmBooktitle{SIGGRAPH Asia 2026 Conference Papers (SA Conference Papers '26), December 01--04, 2026, Kuala Lumpur, Malaysia}
\acmDOI{10.1145/3829340.3842322}
\acmISBN{979-8-4007-2842-6/2026/12}

\AtBeginDocument{%
  }

\usepackage{cleveref}

\input{include/supp_macros}
\input{include/supp_packages}

\begin{document}

\title{ID-V2V: Identity-Preserving Video Restylization}


\author{Yuancheng Xu}
\orcid{0000-0002-2254-5752}
\affiliation{
  \institution{Netflix}
  \country{United States of America}
  \institution{and Eyeline Labs}
  \country{United States of America}
}
\email{xuyuancheng0@gmail.com}

\author{Mingming He}
\authornote{Now at Adobe Nextcam.}
\orcid{0000-0002-9982-7934}
\affiliation{
  \institution{Eyeline Labs}
  \country{United States of America}
}
\email{hmm.lillian@gmail.com}

\author{Pablo Salamanca}
\orcid{0009-0006-0056-0528}
\affiliation{
  \institution{Netflix}
  \country{United States of America}
  \institution{and Eyeline Labs}
  \country{United States of America}
}
\email{Pablosalamanca88@gmail.com}

\author{Li Ma}
\orcid{0000-0002-6992-0089}
\affiliation{
  \institution{Eyeline Labs}
  \country{United States of America}
}
\email{li.ma@scanlinevfx.com}

\author{Yash Kant}
\orcid{0009-0002-8347-4895}
\affiliation{
  \institution{Netflix}
  \country{United States of America}
  \institution{and Eyeline Labs}
  \country{United States of America}
}
\email{ysh.kant@gmail.com}

\author{Emmett Steven}
\orcid{0009-0004-2852-4444}
\affiliation{
  \institution{Netflix}
  \country{United States of America}
}
\email{esteven@netflix.com}

\author{Paul Debevec}
\orcid{0000-0001-7381-2323}
\affiliation{
  \institution{Netflix}
  \country{United States of America}
  \institution{and Eyeline Labs}
  \country{United States of America}
}
\email{debevec@netflix.com}

\author{Ning Yu}
\orcid{0009-0004-6865-1325}
\affiliation{
  \institution{Netflix}
  \country{United States of America}
  \institution{and Eyeline Labs}
  \country{United States of America}
}
\email{ningyu.hust@gmail.com}

\renewcommand{\shortauthors}{Xu et al.}

\input{sections/0_abstract}

\begin{teaserfigure}
  \centering
  \includegraphics[width=0.84\textwidth]{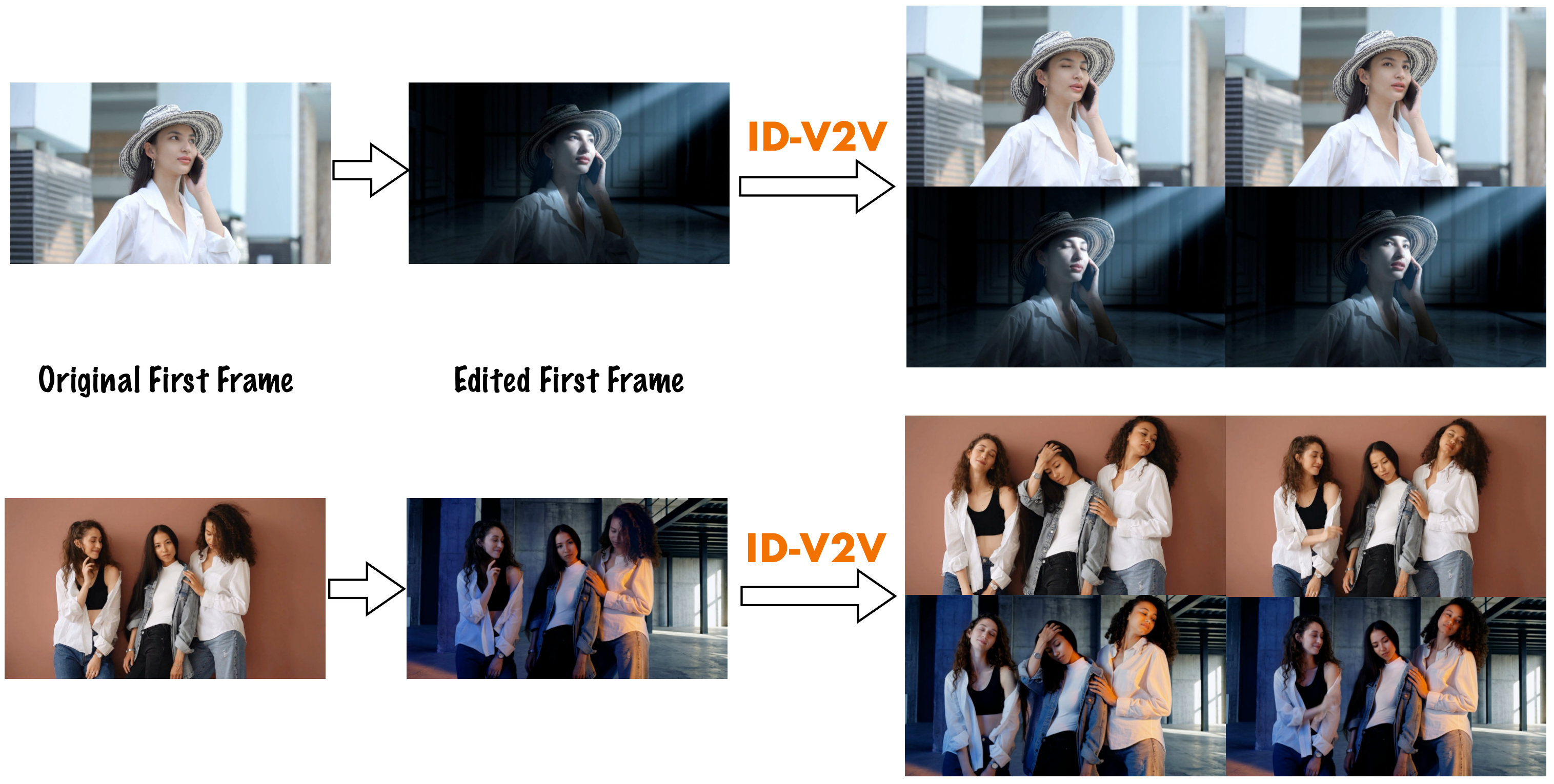}
  \caption{ID-V2V overview. ID-V2V performs identity-preserving video restylization by taking an edited first frame from a source video that specifies new visual styles, such as scene, lighting, and additional visual elements. The model propagates these edits across the entire video while faithfully preserving the human subject’s facial appearance and facial performance, including subtle expressions, eye gaze, and lip synchronization. This enables flexible visual edits without compromising human performance, highlighting the potential of ID-V2V as a human-centric tool for real-world content production.}
  \label{fig:teaser}
\end{teaserfigure}

\maketitle

\input{sections/1_introduction}
\input{sections/2_related_work}
\input{sections/3_method}

\input{sections/4_experiments}

\input{sections/5_conclusion}

\begin{acks}
We would like to thank Lukas Lepicovsky, Mohsen Mousavi, David Rhodes, Nhat Phong Tran, Dacklin Young, and Johnson Thomasson for their production insights. We also thank Jeffrey Shapiro and Ritwik Kumar for their executive support, and Jennifer Lao and Lianette Alnaber for their operational support.
\end{acks}

\clearpage

\input{sections/figure_only}

\clearpage
\bibliographystyle{ACM-Reference-Format}
\bibliography{references}


\end{document}

%% file: include/supp_macros.tex
\usepackage{xcolor}


%% file: include/supp_packages.tex
\usepackage[table]{xcolor}
\usepackage{subcaption}
\usepackage{makecell}  
\usepackage{colortbl}

\usepackage{enumitem}

%% file: sections/0_abstract.tex
\begin{abstract}
In visual storytelling, human performances are central to creative intent and narrative meaning. However, preserving human identity and performance while enabling flexible visual edits remains a fundamental challenge for generative video models. We formalize this challenge as identity-preserving video restylization, a task that propagates scene, lighting, and style changes specified by an edited keyframe across a source video, while strictly preserving facial likeness and performance, including expressions, eye gaze, and lip synchronization.
A key obstacle is the absence of paired training data, as identity-preserving restylized video pairs are rarely available in real-world settings. To address this, we propose a principled decoupling between source-grounded identity preservation and edit-driven video synthesis. Our key insight is that facial appearance and expression should remain invariant, with illumination being the primary permissible variation. We therefore cast identity preservation as a video relighting problem, while modeling visual edit propagation as controlled video synthesis guided by the edited keyframe. Building on this formulation, we introduce ID-V2V, a video-to-video generative framework that integrates complementary control signals: relit facial regions and facial normal maps tightly constrain facial likeness and performance, while edited keyframes and depth sequences enable flexible and temporally coherent generation. This design enables effective construction of training pairs from a single video, eliminating the need for scarce paired data. Extensive experiments demonstrate that ID-V2V significantly outperforms existing methods in preserving facial likeness and fine-grained facial performance, robustly supports both single- and multi-subject scenarios, and delivers high visual quality, highlighting its strong potential as a human-centric tool for real-world content production.
The code is available at: \url{https://github.com/Eyeline-Labs/ID-V2V}.

\end{abstract}

%% file: sections/1_introduction.tex
\section{Introduction}
\label{sec:introduction}

In visual content creation workflows, a highly effective strategy is to capture characters and their performance first, and then design the visual appearance of the video later during post-production. By decoupling performance capture from visual design, this workflow allows creators to preserve character performance—such as facial expressions, motion, and interactions—with high fidelity, while adding or modifying visual elements afterward, including background, lighting, and overall style. This separation also avoids the need to physically construct the final visual setup during capture, reducing constraints from location, set design, and on-set lighting. Together, these properties establish a powerful paradigm centered on faithful preservation of character performance while enabling flexible post-hoc design of the visual presentation.
Meanwhile, recent advances in video generation models~\cite{blattmann2023align, blattmann2023stable,guo2024animatediff,yang2024cogvideox,wan2.1,hacohen2026ltx}, have shown remarkable progress in synthesizing high-quality, temporally coherent videos. 
This naturally raises the question of whether generative video models can enable a similar human-centric paradigm: faithfully preserving character performance—often crucial for creative intent and narrative meaning—while flexibly generating the visual context of the scene.

We formalize this challenge as identity-preserving video restylization, a task that propagates scene, lighting, and appearance changes specified by an edited keyframe across a source video, while preserving facial likeness and performance, including expressions, eye gaze, and lip synchronization. Human subjects may be relit to match the edited scene, while their identity and performance remain unchanged.
The target appearance is specified via an edited keyframe, which provides an intuitive and precise control interface that aligns naturally with existing image-based editing tools used in professional workflows~\cite{wu2025qwen,nanobanana}. This formulation allows productions to capture performances with minimal physical setup and subsequently define the desired visual appearance through keyframe editing, enabling the generation of a consistently restylized video while preserving the original human performance.

However, a central challenge of this task is the lack of large-scale paired training data in real-world settings, as it is difficult to obtain two videos showing the same character performing the identical actions under different visual styles.
To overcome this challenge, our key insight is a principled decoupling between edit-driven synthesis and source-grounded identity preservation. Edit-driven synthesis should remain flexible and creative with respect to the edited keyframe, allowing the model to freely generate the visual content specified by the edit. In contrast, identity preservation must be tightly constrained by the source video. Crucially, we observe that under identity-preserving restylization, facial structure and expression should remain invariant, with illumination constituting the primary source of permissible variation. This observation leads us to formulate identity preservation as a video relighting problem~\cite{debevec2000acquiring}, while instantiating edit-driven synthesis as conditioned image-to-video generation.

Building on this observation, we propose ID-V2V, a video-to-video generative framework that addresses the data scarcity challenge by constructing dedicated control signals from a single training video with distinctive visual style, enabling both edit-driven synthesis and source-grounded identity preservation. Specifically, from the training video we derive relit face regions to simulate changes in facial lighting that may occur between the input video and the stylized output at inference time. We also extract face normal maps as complementary geometric cues. To represent the desired visual edits, we use the first frame of the training video together with a depth sequence extracted from it, which specify the target visual appearance while preserving the underlying scene geometry and motion. Together, these signals allow us to synthesize paired supervision from a single training video, effectively avoiding the need for scarce real-world video pairs for identity-preserving video restylization.

Our experiments on video restylization demonstrate two key advantages of ID-V2V. First, ID-V2V significantly outperforms existing baselines in preserving facial identity and fine-grained facial performance, including expressions and lip synchronization. In contrast, prior methods~\cite{hu2024animate, cheng2025wan, wang2025fantasyportrait, jiang2025vace} typically rely on abstract control signals such as facial landmarks or low-dimensional identity embeddings, which are insufficient to faithfully retain detailed performance cues. Both our quantitative metrics and user study consistently support this finding. Second, ID-V2V reliably supports multi-subject video restylization while maintaining identity preservation for each individual and facial interaction, a capability that remains challenging for existing approaches yet is critical for real-world content creation scenarios.

In summary, our key contributions are as follows:
\begin{enumerate}[topsep=0pt]
    \item We introduce identity-preserving video restylization, a new task motivated by real-world content creation workflows, where visual edits such as background and lighting are specified via an edited keyframe and coherently propagated across a source video while strictly preserving human identity and fine-grained performance.
    \item We propose ID-V2V, a video-to-video generative framework that addresses this task through a principled decoupling between edit-driven synthesis and source-grounded identity preservation. By formulating identity preservation as a video relighting problem and designing complementary control signals, it enables flexible edit propagation while tightly constraining human appearance and performance.
    \item Extensive experiments demonstrate that ID-V2V consistently outperforms existing baselines in preserving facial identity and fine-grained performance, robustly supports multi-subject scenarios, and achieves strong overall visual quality.
\end{enumerate}

%% file: sections/2_related_work.tex
\section{Related work}
\label{sec:related_work}

\paragraph{Identity-Preserving Generation with Reference Images.} 
A substantial body of prior work addresses identity preservation in generative video synthesis by conditioning on reference images or learned identity features, rather than an input performance video. Methods such as ConsisID~\cite{huang2024consistentid}, EchoVideo~\cite{wei2025echovideo}, MagicMirror~\cite{zhang2025magicmirror}, ID-Animator~\cite{he2024id}, Phantom~\cite{liu2025phantom}, HuMo~\cite{chen2025humo}, and VirtuallyBeing~\cite{xu2025virtually}
, IRPO~\cite{shen2026identity}, Stand-In~\cite{xue2026standin}, and IP-FaceDiff~\cite{anand2025ipfacediff}, primarily operate in text-to-video or image-to-video settings, where one or a few reference images are used to inject identity cues into the generation process.
While these approaches improve visual identity consistency, they do not take a source video as input and thus cannot exploit temporal cues from real performances, such as subtle facial expressions and performances. As a result, identity preservation is often limited to coarse appearance, and alignment with a captured performance remains underconstrained. They also typically lack artist-driven editing interfaces, such as edited keyframes, limiting fine-grained and controllable restylization in practical production settings.

\paragraph{Video-to-Video Generation and Performance Transfer.}
Several video-to-video generation methods~\cite{burgert2025go, jiang2025vace} aim to transfer motion or structural information from an input video through intermediate control signals. Approaches such as Wan-Animate~\cite{cheng2025wan}, AnimateAnyone~\cite{hu2024animate}, VACE~\cite{jiang2025vace}, FantasyPortrait~\cite{wang2025fantasyportrait}, and SteadyDancer~\cite{zhang2025steadydancer} infer facial identity primarily from a single input frame and use facial or motion control signals to guide video generation. Because identity cues are not grounded in the full source video, these methods struggle with multi-view identity preservation when the viewing direction changes, or must implicitly infer identity under variations in expression and lighting, which can lead to inaccurate identity reconstruction. In addition, facial performance is typically driven by abstract or low-dimensional representations—such as facial pose or landmarks (e.g., VACE, AnimateAnyone, SteadyDancer) or compact facial and expression embeddings (e.g., FantasyPortrait, Wan-Animate)—rather than directly leveraging the source video. Consequently, while these approaches can generate temporally coherent facial motion, they often fail to faithfully preserve fine-grained facial performance details, including subtle expressions, eye gaze, and lip synchronization, which are critical for identity-preserving video restylization. More broadly, these methods are not explicitly designed for this setting, as they prioritize motion or structural transfer over strict identity and performance fidelity.

\paragraph{General Stylization.}
General image and video stylization transforms the visual appearance of the input according to an artistic reference or edited exemplar, enabling effects such as cartoonization, oil-painting styles, or other user-specified appearances~\cite{gatys2016image,johnson2016perceptual,huang2017arbitrary,ruder2016artistic,jamrivska2019stylizing,spetlik2025structureiser}. Such methods may stylize the face or the entire human subject, potentially altering identity-defining appearance and fine-grained performance details. In contrast, identity-preserving video restylization imposes a stricter constraint: the subject may only be relit, while their intrinsic identity and performance must be faithfully preserved.

\paragraph{Relighting and Illumination Control.} 
Relighting is an active area of research that aims to manipulate illumination conditions while preserving the underlying visual content. A substantial body of work studies single-image portrait relighting and facial illumination manipulation, focusing on modifying facial lighting while preserving human identity~\cite{he2024diffrelight,mei2025lux,rao2024_lite2relight,ponglertnapakorn2023difareli,mei2024holo_cvpr}. 
Related methods also address portrait relighting for background replacement, lighting harmonization, and shadow removal~\cite{pandey2021total,ren2024relightful,yoon2024generative,kim2024switchlight}.
These approaches leverage human face priors, real captured data, or physically based rendering models, but are primarily designed for single images and can exhibit temporal inconsistencies when applied to video. Recently, more works have been proposed to address relighting and illumination editing in videos, with an emphasis on temporal consistency across frames. Representative methods include Light-X~\cite{liu2025light}, Lumen~\cite{zeng2025lumen}, Light-A-Video~\cite{zhou2025lightavideo}, RelightVide~\cite{fang2025relightvid}, RelightMaster~\cite{bian2025relightmaster} and UniRelight~\cite{he2025unirelight}. 
These methods modify video lighting while maintaining temporal coherence. However, they mainly rely on text-based or parametric controls and do not support edited keyframes as explicit user input. They also do not target identity-preserving restylization, where identity and performance must be preserved under substantial visual changes beyond lighting. We therefore leverage portrait image relighting as a control signal that changes lighting while preserving facial features and identity.

%% file: sections/3_method.tex
\section{Method}
\label{sec:method}

\subsection{Task Definition}

\paragraph{Task Inputs and Output.}
We study keyframe-based identity preserving video restylization, illustrated in~\ref{fig:teaser}. The task takes two primary inputs. (1) A \textbf{source video} that contains a human performance. This video may be captured in a simple or different environment from the desired final output, but the goal is to faithfully preserve the character’s identity, including both facial likeness and dynamic facial performance throughout the sequence. (2) An \textbf{edited first frame}, obtained by modifying the first frame of the source video to specify the desired visual edits, such as changes to the environment, lighting, or other visual elements. These edits can be created manually by artists or generated using existing image editing models, providing an intuitive interface for visual control. Given these inputs, the task is to propagate the visual edits specified by the keyframe across the entire video while strictly preserving the character’s facial likeness and performance in the generated video.

\paragraph{Human-centric Content Creation.} 
By enabling intuitive first-frame edits that automatically propagate across the entire video, this formulation provides a convenient interface for creators to redesign scenes, environments, lighting, and other visual elements after the original performance has been captured. At the same time, it treats the human subject and their performance as invariant, faithfully preserving facial identity and expressive dynamics throughout the video. This aligns with a fundamental requirement in many forms of content creation, where human performance anchors storytelling and artistic intent.

\paragraph{Challenge in Identity Preservation.} 
Preserving facial likeness and fine grained facial performance under video restylization is inherently difficult, because these attributes are encoded in subtle cues such as lip synchronization, micro expressions, eye gaze, and emotional dynamics, which can be easily disrupted by visual changes. This challenge arises from both modeling and data limitations. (1) Modeling challenge. Such subtle performance cues are poorly captured by abstract representations such as depth maps or identity embeddings, causing existing methods to degrade identity and expression fidelity. (2) Data challenge. Large scale paired training data for this task is scarce, as it is difficult to obtain two videos showing the same character performing identical actions under different visual styles.

\subsection{Decoupling Edit-Driven Synthesis and Identity Preservation}

\begin{figure} [t!]
    \centering 
    \includegraphics[width=0.98\columnwidth]{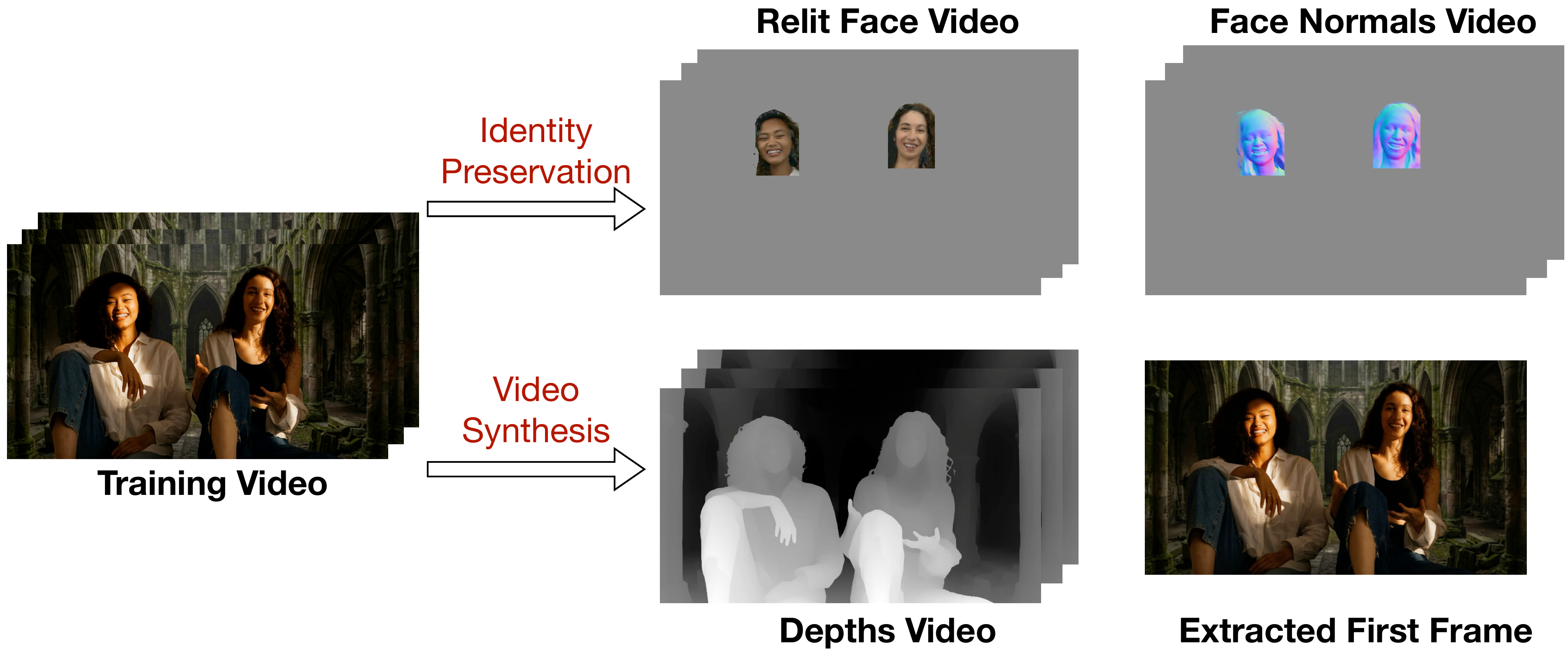}
    \caption{
    Training data construction. Identity-preserving video restylization is decomposed into video synthesis and identity preservation. For synthesis, the first frame and a depth sequence are extracted. For identity preservation, a relighting model converts the cinematic lighting of the training video faces into regular lighting that simulates the lighting of the source video provided at inference time, while face normals provide additional geometric cues. This design allows the model to learn to transform regular facial lighting into cinematic lighting while preserving likeness and subtle expressions.
    }
    \label{fig:trainingConditions}
\end{figure}

\paragraph{Overview.} 
Our key insight is to decouple edit-driven synthesis from source-grounded identity preservation. The synthesis component determines what visual content to generate and is guided by the edited keyframe, allowing flexible generation of the desired scene appearance. In contrast, identity preservation specifies what must remain unchanged: the human subject is tightly anchored to the source video to faithfully retain facial likeness and fine grained facial performance. Guided by this separation, ID V2V employs two complementary types of control signals, illustrated in Figure~\ref{fig:trainingConditions}. Relit face video and face normal maps enforce identity preservation, while the first frame and depth sequence guide video synthesis. This design enables the construction of effective training pairs, thereby overcoming the scarcity of real-world before-and-after restylization examples.

\paragraph{Edit-Driven Synthesis. } 
This component translates keyframe edits into a temporally coherent video. We implement it as image to video generation with depth control. The model takes the edited keyframe together with a depth sequence extracted from the source video using DepthAnything 2~\cite{yang2024depth}. The edited keyframe provides explicit guidance on the desired appearance and stylistic changes, while the depth signal encodes global motion and scene layout. Together, these signals enable consistent propagation of complex visual edits across time while preserving motion coherence and scene geometry.

\paragraph{Identity Preservation as Relighting.} 
Preserving human identity and fine grained facial performance is particularly challenging in video restylization. These attributes are encoded in subtle, high frequency visual cues that are poorly captured by abstract control signals such as depth maps or facial landmarks. Instead, faithful preservation requires directly leveraging the pixel level information of human regions in the source video. A key observation is that identity preserving restylization should keep facial structure and expression unchanged, while illumination accounts for most permissible variation. 
Therefore, we formulate identity preservation as a video relighting problem, where the lighting may change while the underlying facial identity and performance remain fixed.

\paragraph{Relit Face Control. }
Following the relighting formulation for identity preservation, we condition the model on relit face and head regions to explicitly model lighting changes on the human subject. Generating such relit data requires an  accurate relighting model that can modify facial illumination while preserving fine details of the face and head with high fidelity. To meet this requirement, we train a dedicated image based face relighting model. We then apply this model frame by frame to the detected face and head regions of the training video, producing relit face sequences that serve as conditioning signals during training. This design effectively adapts image based relighting as a control signal for video restylization, providing a principled mechanism for preserving identity.

\paragraph{Relighting Model Design. }
The design of our relighting model is inspired by the image de-lighting approach of LuxPostFacto~\cite{mei2025lux}. However, unlike the flat lit normalization used in LuxPostFacto, we relight faces from the training video to a regular lighting condition. This choice is motivated by the training setup of ID-V2V: the relit faces serve as inputs during training and therefore should resemble the lighting conditions typically encountered at inference time, where source videos are often captured under ordinary  lighting. Meanwhile, faces in the training videos frequently exhibit cinematic or stylized illumination, which we treat as the desired target appearance. As a result, ID-V2V learns to transform regular facial illumination into cinematic or stylized lighting for the human regions. In practice, we simulate regular illumination using a top lighting configuration. As shown in \cref{fig:relighting_demo}, the relighting model removes the original illumination and applies consistent regular lighting to the face regions, producing reliable conditioning signals for facial likeness and performance.

\paragraph{Relighting Model Training. }
To train the relighting model, we construct a large scale dataset of paired face images using the One Light at a Time (OLAT)~\cite{debevec2000acquiring} data collected in LuxPostFacto~\cite{debevec2000acquiring}. For each pair, the input image is a face rendered under arbitrary environmental illumination, synthesized by compositing OLAT captures with HDR environment maps randomly sampled from PolyHeaven~\cite{polyhaven}, following the image relighting procedure of~\cite{debevec2000acquiring}. The target image is the same face relit with a vertically half white, half black HDRI that produces the desired top lighting configuration. Using data from 67 participants with diverse head poses, facial expressions, and camera viewpoints, this process yields approximately 330,000 image pairs. The model architecture and training settings largely follow the image relighting model of LuxPostFacto~\cite{mei2025lux}. Specifically, we adopt the pretrained SVD model~\cite{blattmann2023stable} as the base network and reduce the number of frames to one to convert it into an image based model. We further modify the model to condition on an HDR map and fine tune it on the paired training images for 200K steps. During inference, we detect and crop face regions from each video frame and apply the relighting model independently to produce their top lit versions.

\paragraph{Face Normals Control. }
While relit faces provide rich cues about facial appearance under varying illumination, they remain insufficient to fully constrain facial geometry due to the inherent ambiguity between lighting and surface shape, which can lead to geometric drift and identity degradation. To explicitly encode facial structure and stabilize fine-grained identity cues, we additionally condition on surface normal maps predicted using the DAViD~\cite{saleh2025david} model. Relighting enforces appearance consistency, while normals provide complementary geometric constraints on facial shape. Together, they form a complete and robust control signal for faithful preservation of facial identity and performance under stylistic changes.

\subsection{Architecture, Training and Inference}

\begin{figure} [t!]
    \centering 
    \includegraphics[width=0.8 \columnwidth]{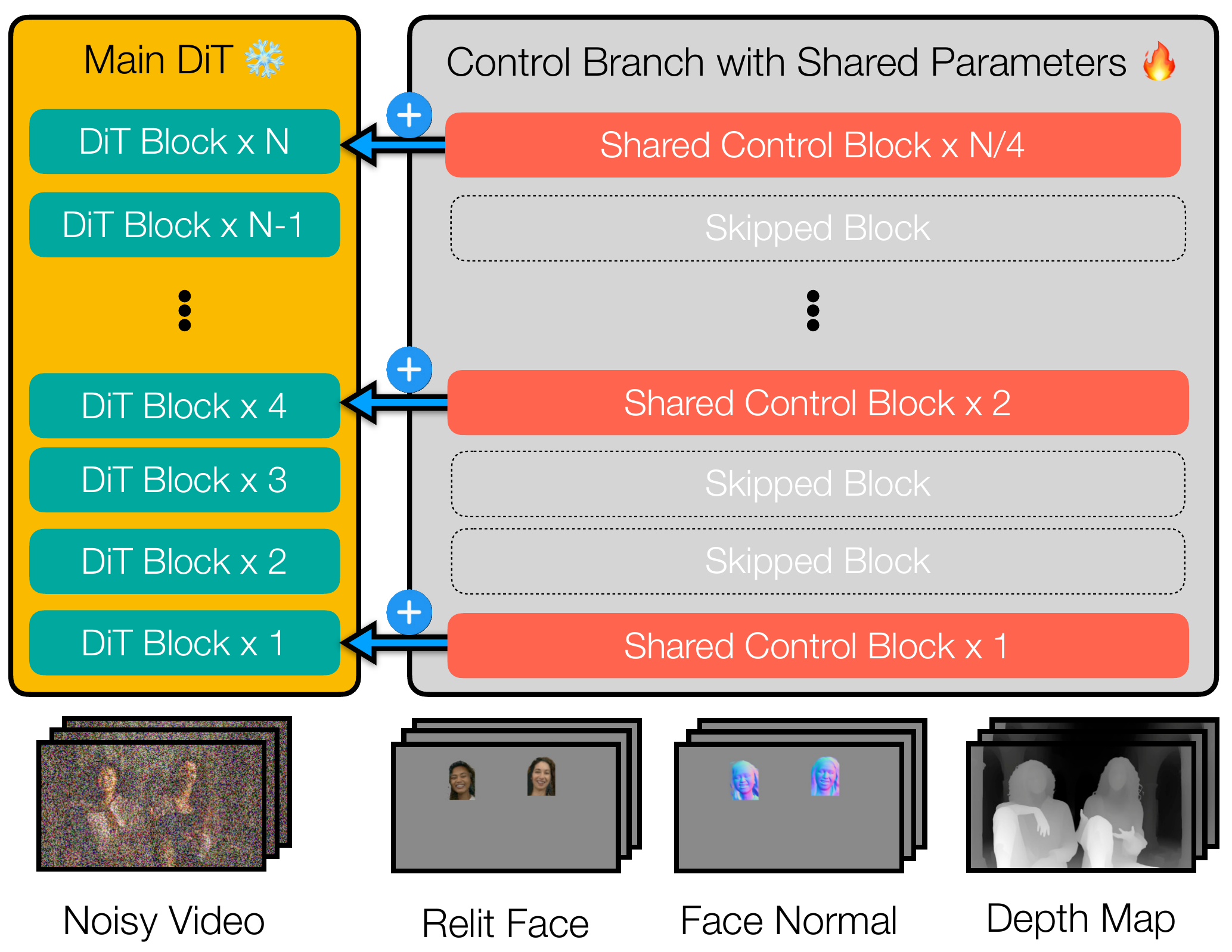}
    \caption{ID-V2V architecture. The model conditions on relit facial regions and facial normal maps to preserve facial appearance and performance, and on depth sequences together with the edited first frame and text prompts (not shown) to guide edit-conditioned video synthesis. The relit face, face normal, and depth controls share the same ControlNet parameters initialized by VACE~\cite{jiang2025vace}, and their influences are fused via summation before being injected into the main DiT blocks.}
    \label{fig:pipeline}
\end{figure}

\paragraph{Model Architecture and Training. }
To support multiple spatially aligned control videos, we build upon VACE~\cite{jiang2025vace} based on Wan2.1~\cite{wan2.1}, which natively supports video conditioning, and extend it to a multi ControlNet~\cite{zhang2023adding} architecture (Figure~\ref{fig:pipeline}). To reduce computational overhead, each control signal is processed using the same VACE context block, and the resulting features are fused by summation before being injected into the main DiT blocks. For keyframe control, we follow the VACE input format and replace the first frame in the control videos with the edited first frame. The model also conditions on a text prompt and is trained jointly with all control signals.

\paragraph{Control Signals During Inference.}
During inference, we directly use the original facial region pixels from the source video to avoid any loss of identity information. This follows the training design, where relit faces are used only during training to simulate the lighting conditions of the source video at inference. As a result, the model learns to transform the regular facial lighting of the input source video into the stylized illumination specified by the edited first frame. Although the model is trained with pixel and normal guidance restricted to the facial region, the guided region can be expanded at inference time. Providing full-body pixels and normals helps preserve the appearance and structure of the entire body, while extending the guidance to the full frame can further enable full-scene relighting to some extent. The global depth video is kept the same as during training to maintain geometric consistency and coherent propagation of the visual edits across the sequence.

%% file: sections/4_experiments.tex
\section{Experiments}
\label{sec:experiments}

\subsection{Experimental Setup}

\paragraph{Training. }
We collect a dataset of 40k human-centric videos featuring both single- and multi-subject scenarios, covering diverse scenes, lighting, actions, and facial expressions. To support the post-training requirements of VACE, we generate textual descriptions for each video using the QwenVL2.5-32B~\cite{bai2025qwen2} model. For identity preservation, we run SCRFD face detection~\cite{guo2021sample} on all videos and retain up to five faces per frame, selected by pixel area, to accommodate multi-subject content commonly encountered in real-world content creation. As a result, the relit-face and normal-based control signals contain at most five faces per frame, even when additional faces are present. All models are trained at a resolution of 1280 × 720 with 81 frames per video. To ensure effective learning of identity-related cues, we discard videos in which more than 40\% of frames contain no detected faces.

\begin{figure*}[ht]
    \centering 
    \includegraphics[width=0.93\textwidth]{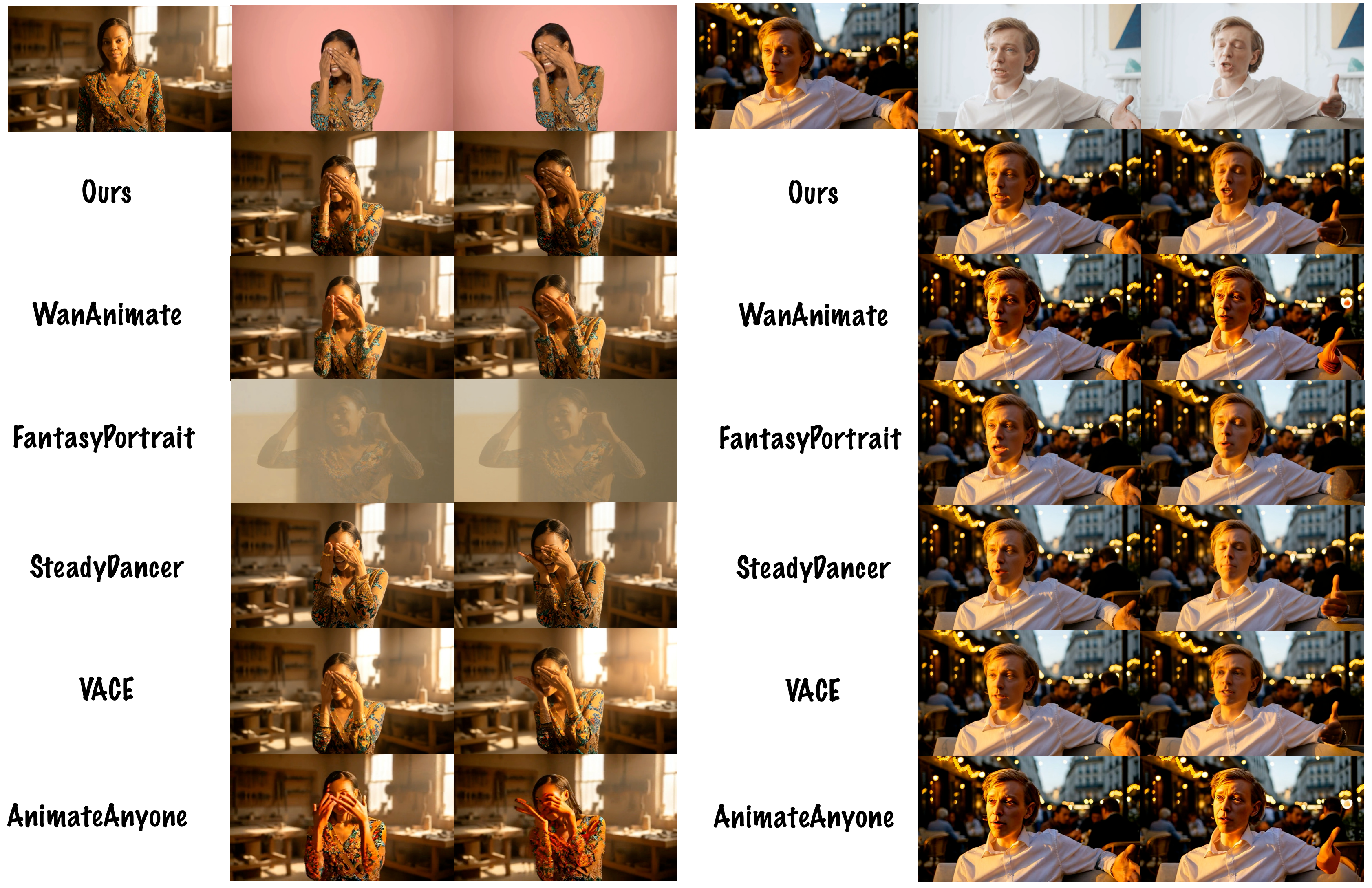}
    \caption{
   Single-subject comparison. The first row shows the edited first frame and the source video, while the following rows present the generated videos. ID-V2V achieves the strongest facial identity preservation and faithfully retains facial performance, and can handle occlusion around the face. 
    }
    \label{fig:single_subject_baseline}
\end{figure*}

\paragraph{Evaluation Datasets. }
We construct two human-centric evaluation datasets: 100 single-subject videos and 60 two-subject videos, covering diverse actions and facial expressions. We report results separately on these two settings. For quantitative evaluation, we restrict the maximum number of subjects to two, as existing baselines are unable to reliably handle more subjects; nevertheless, ID-V2V naturally supports multi-subject scenarios, which we additionally demonstrate qualitatively.

To specify the target style, we first use Gemini-2.5~\cite{comanici2025gemini} to analyze the original first frame and propose a target background and lighting configuration. These proposed edits are then executed by an image editing model, Qwen-Image-Edit~\cite{wu2025qwen}, with depth and Canny edge conditions, producing substantial changes in background and illumination. We further perform manual filtering to ensure high-quality edited keyframes. Each evaluation sample therefore consists of a source video and an edited first frame, and we assess how well different methods propagate the specified edits across the full video while preserving human facial likeness and performance.

\paragraph{Baselines. }
We select baseline methods that take an edited first frame as input and provide facial or motion control to guide video generation. Based on this criterion, we compare against AnimateAnyone~\cite{hu2024animate}, a UNet-based method, as well as several DiT-based approaches built on Wan model's backbones, including VACE~\cite{jiang2025vace}, FantasyPortrait~\cite{wang2025fantasyportrait}, and SteadyDancer~\cite{zhang2025steadydancer}, and WanAnimate~\cite{cheng2025wan}.

\subsection{Evaluation Metrics}

\paragraph{Facial Likeness. }
To evaluate facial likeness between the generated and source video, we apply SCRFD face detection~\cite{guo2021sample} to extract face crops and compute identity similarity using AdaFace~\cite{kim2022adaface}. Frames without detected faces are discarded. For videos with multiple subjects, we resolve face correspondence across videos using AdaFace similarity scores to consistently match each generated face with its corresponding reference identity.

\paragraph{Facial Performance Preservation. } We measure the similarity of facial expressions using two widely adopted expression representations. The first is Action Units (AUs) from the Facial Action Coding System (FACS)~\cite{ekman1978facial}, which capture fine-grained facial muscle activations. The second is expression vectors extracted from EMOCA~\cite{danvevcek2022emoca}, an emotion-aware 3D face reconstruction model that estimates detailed facial expression parameters. For each face in the generated video, we compute the cosine similarity between its expression representation (either AU vectors or EMOCA expression vectors) and that of the corresponding face in the source video, denoted as Exp-AU and Exp-EMOCA, respectively. These metrics provide complementary evaluations of facial expression preservation.

\subsection{Results}

\begin{table*}[htp!]
    \centering
    \caption{Quantitative results for identity-preserving video restylization. User study columns report the percentage of responses favoring each method for each evaluation criterion. $\Uparrow$/$\Downarrow$ indicates a higher/lower value is better. \textbf{Bold} indicates the best results. “Ours” denotes the full ID-V2V model; “Ours w/o face normals” and “Ours w/o face video” denote ablation variants. }
    
    \resizebox{0.97\linewidth}{!}{
    \begin{tabular}{l|c|c|c|c|c|c|c|c|c|c|c|c}
        \toprule
        \textbf{Method}
        & \multicolumn{3}{c|}{\textbf{Facial Metrics} $\Uparrow$}
        & \multicolumn{2}{c|}{}
        & \multicolumn{4}{c|}{\textbf{VBench \& VBench2.0} $\Uparrow$}
        & \multicolumn{3}{c}{\textbf{UserStudy} $\Uparrow$} \\
        
        & \makecell{AdaFace}
        & \makecell{Exp- \\AU}
        & \makecell{Exp- \\EMOCA}
        & \makecell{CLIP-T $\Uparrow$}
        & \makecell{CLIP-I $\Uparrow$}
        & \makecell{Subject\\Consistency}
        & \makecell{Background\\Consistency}
        & \makecell{Temporal\\Flickering}
        & \makecell{Human\\Anatomy}
        & \makecell{Facial\\Likeness}
        & \makecell{Facial Perf.\\Preservation}
        & \makecell{Video\\Quality} \\
        \midrule

        \multicolumn{13}{c}{\textbf{Single-subject video restylization}} \\
        \midrule
        AnimateAnyone   & 0.402 & 0.477 & 0.690 & 0.331 & 0.976 &  0.961 & 0.951 & 0.971 & 0.923 & 0.23\% & 0.23\% & 0.23\% \\
        FantasyPortrait & 0.507 & 0.550 & 0.767 & 0.332 & \textbf{0.997} &  0.953 & 0.950 & 0.991 & 0.930 & 1.82\% & 2.27\% & 2.28\%\\
        VACE            & 0.465 & 0.580 & 0.776 & \textbf{0.334} & 0.992 & 0.939 & 0.932 & 0.985 & 0.923 & 7.95\% & 8.18\% &  10.05\% \\
        WanAnimate      & 0.529 & 0.600 & 0.841 & \textbf{0.344} & 0.993 & \textbf{0.962} & 0.945 & 0.987 & 0.941 & 10.45\% &  11.36\% &  12.56\% \\
        SteadyDancer    & 0.479 & 0.554 & 0.745 & 0.328 & 0.993 & 0.955 & 0.945 & 0.985 & 0.931 & 4.55\% &  3.64\% & 4.79\% \\
        Ours            & \textbf{0.701} & \textbf{0.668} & \textbf{0.899} & \textbf{0.344} & 0.993 & 0.960 & \textbf{0.954} & \textbf{0.993} & \textbf{0.945} & \textbf{75.00\%} & \textbf{74.32\%} &  \textbf{70.09\%} \\
        Ours (w/o face normals)            & 0.676 & 0.638 & 0.881 & 0.343 & 0.992 & 0.958 & 0.948 & 0.990 & 0.943 & - & - & - \\
        Ours (w/o face video)            & 0.574 & 0.585 & 0.798 &  \textbf{0.344} & 0.993 & 0.955 & 0.952 & 0.992 & 0.938 & - & - & - \\
        \midrule

        \multicolumn{13}{c}{\textbf{Two-subject video restylization}} \\
        \midrule
        FantasyPortrait & 0.433 & 0.624 & 0.747 & \textbf{0.331} & \textbf{0.997} & 0.930 & 0.929 & 0.986 & 0.885 & 0.65\% & 0.22\% & 1.31\% \\
        VACE            & 0.401 & 0.639 & 0.796 & 0.329 & 0.994 & 0.956 & 0.947 & 0.987 & 0.876 & 9.80\% & 10.00\% & 12.42\% \\
        WanAnimate      & 0.432 & 0.617 & 0.732 & 0.330 & 0.995 & 0.975 & 0.954 & \textbf{0.988} & 0.921 & 3.92\% & 3.70\% & 5.23\% \\
        Ours            & \textbf{0.637} & \textbf{0.709} & \textbf{0.901} & \textbf{0.331} & 0.996 & \textbf{0.979} & \textbf{0.956} & \textbf{0.988} & \textbf{0.926} & \textbf{85.62\%} & \textbf{86.09\%} &  \textbf{81.05\%} \\
        \bottomrule
    \end{tabular}
    }
    \label{tab:main}
\end{table*}

\paragraph{General Video Quality. }
We evaluate general video quality along three complementary aspects: prompt alignment, temporal consistency, and overall visual fidelity. Prompt alignment is assessed using CLIP-T \cite{radford2021learning} by computing the average CLIP similarity between the input prompt and generated frames. Temporal consistency is measured via the average CLIP image similarity between consecutive frames. To further assess video fidelity and perceptual quality, we benchmark our method using VBench \cite{huang2024vbench}, including aesthetic quality, imaging quality, subject consistency, background consistency, temporal style, and VBench-2.0 \cite{zheng2025vbench} to evaluate human anatomy correctness.

\paragraph{User Study. }
We conduct a user study on both the single-subject and two-subject evaluation datasets, each containing 20 samples, with 26 participants to compare video restylization methods. For each example, participants are shown the source video and the edited first frame, followed by videos generated by different methods, and are asked to select the best result under each criterion. The study includes three evaluation criteria: (i) Identity Similarity, which method makes the generated person look most like the source person; (ii) Expression Preservation, which method best preserves facial performance, including lip synchronization, eye-gaze direction, and subtle expressions and emotions; and (iii) Visual Quality, which method produces the best overall visual quality.

\paragraph{Qualitative Results.}
Visual comparisons between ID-V2V and baseline methods are shown in \cref{fig:single_subject_baseline} for the single-subject case and in \cref{fig:two_subject_baseline} for the two-subject case, with frame-by-frame comparisons. In the single-subject setting, ID-V2V preserves facial likeness more faithfully than the baselines, retaining fine-grained details such as teeth shape, subtle facial muscle movements (e.g., eyebrow motion during speech), and eye blinks, at a level of fidelity not achieved by competing methods. Notably, ID-V2V also handles face occlusion effectively, as illustrated in the left example of \cref{fig:single_subject_baseline}, maintaining consistent facial performance before and after the occlusion. In contrast, FantasyPortrait produces corrupted results in occlusion scenarios. In the two-subject setting, methods that rely on compact facial or expression representations (e.g., WanAnimate and FantasyPortrait) struggle to preserve both facial identity and expression, with degradation becoming more pronounced when multiple subjects are present. In contrast, ID-V2V better preserves individual identities and facial performances, which in turn allows it to maintain the interactions between subjects in the source video.

\paragraph{Quantitative Results on Identity Preservation.}
Quantitative results are reported in~\cref{tab:main}. We evaluate facial identity preservation using the AdaFace score and assess facial performance preservation using Exp-AU and Exp-EMOCA. Across both the single-subject and two-subject settings, ID-V2V consistently outperforms all baseline methods, highlighting its ability to preserve both facial identity and fine-grained facial performance under video restylization.

\paragraph{Quantitative Results on General Video Quality.}
As shown in \cref{tab:main}, ID-V2V achieves superior or, in most cases, comparable performance to baseline methods on general video quality metrics. This indicates that training the model for identity-preserving video restylization does not compromise its overall video generation capability. Importantly, baseline methods that are primarily designed for face-specific generation (e.g., FantasyPortrait and WanAnimate) exhibit noticeable degradation in general video quality when moving from single-subject to two-subject scenarios. In contrast, ID-V2V maintains strong performance in both settings, underscoring its robustness and effectiveness for multi-subject video generation.

\paragraph{User Study.}
User study results are reported in \cref{tab:main}. Across all evaluation criteria—including facial appearance preservation, facial performance preservation, and overall visual quality—ID-V2V is consistently preferred over all baseline methods. In the single-subject setting, ID-V2V achieves win rates exceeding 74\% on both facial likeness and facial performance, and over 70\% on overall video quality. The advantage becomes even more pronounced in the two-subject setting, where ID-V2V attains over 85\% win rates on facial likeness and performance, and more than 80\% on overall video quality. These results show that ID-V2V more effectively preserves identity and facial performance with high-quality video generation, particularly in challenging two-subject scenarios.

\paragraph{More Subjects.}
Results demonstrating ID-V2V on videos with more than two subjects are shown in \cref{fig:more_subject}. ID-V2V consistently preserves each subject’s facial appearance and expressions, as well as the interactions between subjects, while restylizing the video into a new scene with different visual elements. These results indicate that ID-V2V effectively supports multi-subject video restylization, where maintaining inter-subject facial performance and interaction is often critical for conveying narrative and semantic content in real-world applications.

\paragraph{Ablation Study on Facial Conditions.}
We ablate the two facial conditions in ID-V2V: relit face videos and face normals. Specifically, we train two variants, one without face normals and one without relit face videos. Quantitative results are reported in~\cref{tab:main}. Removing the relit face video condition causes a significant drop in facial preservation metrics, showing that directly leveraging facial pixels is crucial for maintaining fine-grained facial details. Removing face normals also degrades facial metrics, although to a lesser extent, suggesting that geometric facial cues provide complementary guidance. Together, these results show that training with relit face videos and face normals jointly achieves strong identity and fine-grained facial performance preservation.

\begin{figure}[t!]
    \centering
    \includegraphics[width=0.95\columnwidth]{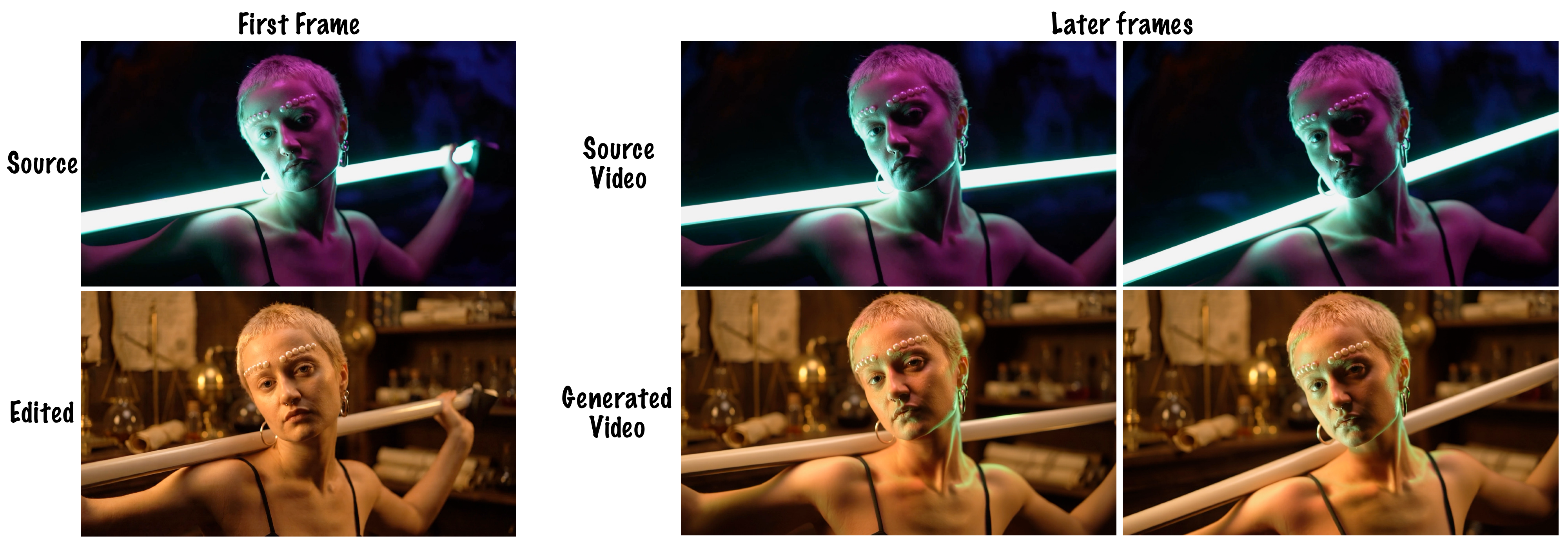}
    \caption{Failure case under irregular source lighting. Although the edited first frame removes the strong green illumination present in the source video, an unintended green color cast reappears in later generated frames, resulting in inconsistent lighting over time.}
    \label{fig:fail_extreme_light}
\end{figure}

\paragraph{Failure mode: extreme source lighting.}
A failure mode occurs under extreme or irregular source lighting, such as strong colored illumination or hard shadows. As shown in~\cref{fig:fail_extreme_light}, although the edited first frame removes the green illumination, residual lighting cues reappear in later generated frames, resulting in an unintended color cast and temporally inconsistent appearance. For more reliable results, we recommend using source videos captured under relatively regular lighting, which better matches the conditions represented in the training data.

\paragraph{Relighting. }
Although ID-V2V is trained using facial relighting supervision, it can be naturally extended to full-scene relighting at inference time. During inference, instead of providing only the facial region as the relighting guidance, we directly use the entire source video as input, allowing the model to exploit appearance cues from the full scene. As shown in~\cref{fig:relighting_generation}, ID-V2V not only relights the face but also consistently relights the body, surrounding objects, and the background. Despite never being explicitly trained for full-scene relighting, the learned facial relighting capability generalizes well beyond the face, producing coherent illumination changes across the entire scene.

%% file: sections/5_conclusion.tex
\section{Conclusion and Limitations}
\label{sec:conclusion}

In this paper, we address a fundamental challenge in generative video for content production: preserving faithful human identity and performance, both of which are central to creative intent and narrative meaning, while still enabling flexible visual edits to the scene. We formalize this challenge as identity-preserving video restylization, a human-centric task that propagates visual edits specified by an edited keyframe across a source video while strictly maintaining facial likeness and fine-grained facial performance such as expressions, gaze, and subtle motion dynamics. Unlike conventional video editing or style transfer, the goal is not merely visual consistency but the faithful preservation of a person’s identity and acting performance from a source video. Crucially, real production settings frequently involve multiple interacting subjects, where maintaining each individual’s identity as well as the temporal consistency of their facial interactions is essential for preserving semantic meaning and narrative coherence. To overcome the lack of paired training data inherent to this task, we propose ID-V2V, a video-to-video framework that decouples source-grounded identity preservation from edit-driven video synthesis, enabling the effective construction of training pairs from a single video without requiring explicit paired supervision. This decoupling is realized through complementary control signals: relit facial regions and facial normal maps tightly constrain identity and performance, while edited keyframes and depth sequences guide flexible, spatially consistent, and temporally coherent generation. Extensive experiments demonstrate that ID-V2V substantially outperforms existing methods in preserving facial identity, subtle performance cues, and inter-subject interactions, while robustly supporting multi-subject scenarios and maintaining high overall video quality. Together, these results highlight the promise of ID-V2V as a practical and scalable human-centric approach for real-world content production.

\paragraph{Limitations. }
Our method has two main limitations. First, because the source depth sequence provides guidance for scene structure and motion, large geometric discrepancies between the source video and the edited keyframe can make this conditioning overly restrictive. Moderate background changes are generally tolerated, but extreme differences in scene layout or depth may conflict with the source geometry and reduce edit fidelity. This limitation could be mitigated by training with depth dropout, allowing the model to optionally operate without depth guidance at inference time when the target scene differs substantially from the source. Second, the method is designed for source videos captured under relatively regular lighting, consistent with both its training conditions and the intended capture-first, restyle-later workflow. Under extreme or irregular illumination, such as strong colored lighting or hard shadows, residual source-lighting cues may propagate into the generated video, producing unintended color casts or temporally inconsistent appearance.

%% file: sections/figure_only.tex
\begin{figure*}[ht]
    \centering 
    \includegraphics[width=0.95\textwidth]{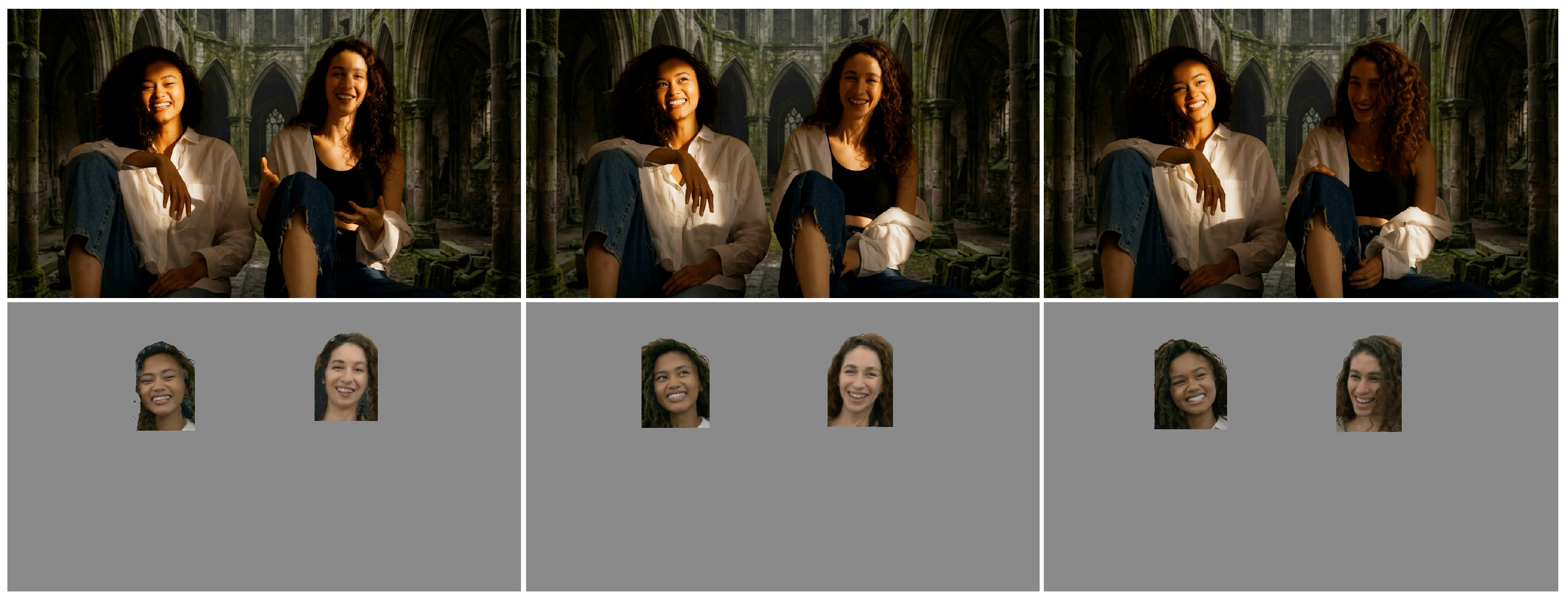}
    \caption{\textbf{Top}: Training video with cinematic lighting. \textbf{Bottom}: Relit face video produced by our image based relighting model. The original facial illumination is removed and replaced with regular lighting to simulate the lighting conditions encountered during ID-V2V inference, while preserving facial likeness and subtle expressions.
    }
    \label{fig:relighting_demo}
\end{figure*}

\begin{figure*}[h]
    \centering 
    \includegraphics[width=0.99\textwidth]{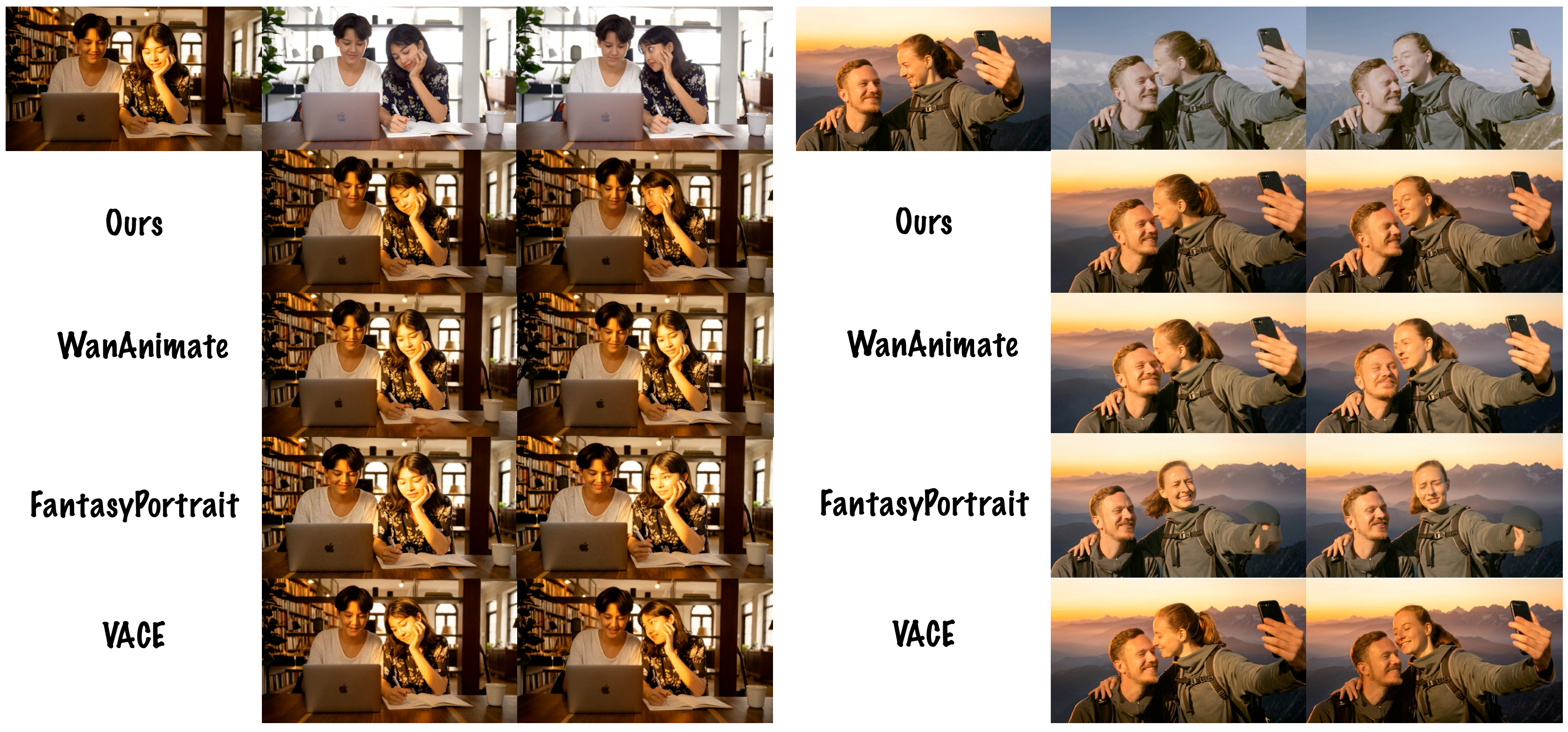}
    \caption{
   Two-subject comparison. The first row shows the edited first frame and the source video, while the following rows present the generated videos. ID-V2V achieves the strongest facial identity preservation and faithfully retains facial performance, as well as subjects' facial interaction. 
    }
    \label{fig:two_subject_baseline}
\end{figure*}

\begin{figure*}[ht]
    \centering 
    \includegraphics[width=0.95\textwidth]{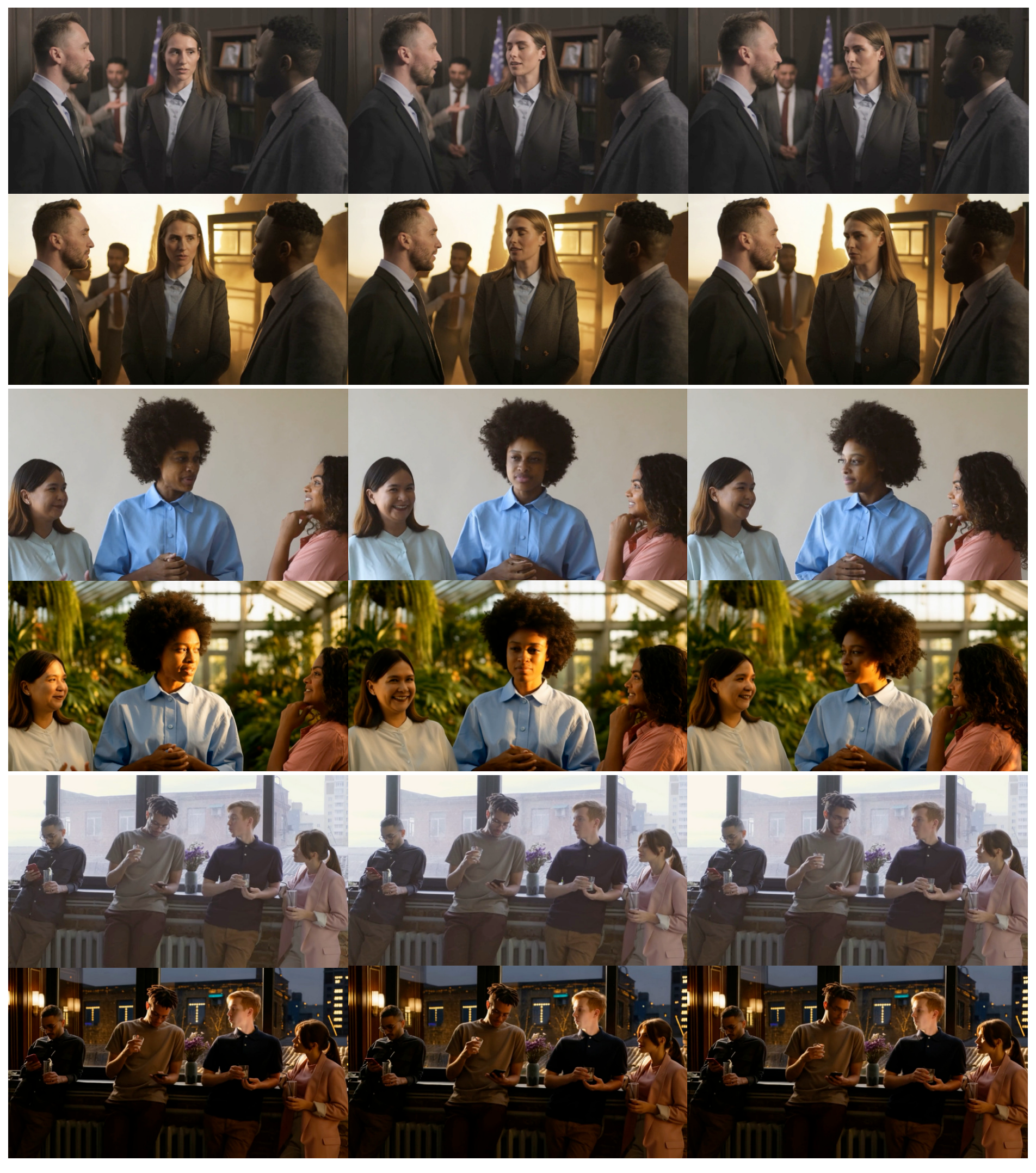}
    \caption{
    More than two subjects. In each pair, the first row is the source video and the second row is generated by ID-V2V, which reliably handles video restylization while preserving both individual identities and interactions among multiple subjects.
    }
    \label{fig:more_subject}
\end{figure*}

\begin{figure*}[ht]
    \centering 
    \includegraphics[width=0.95\textwidth]{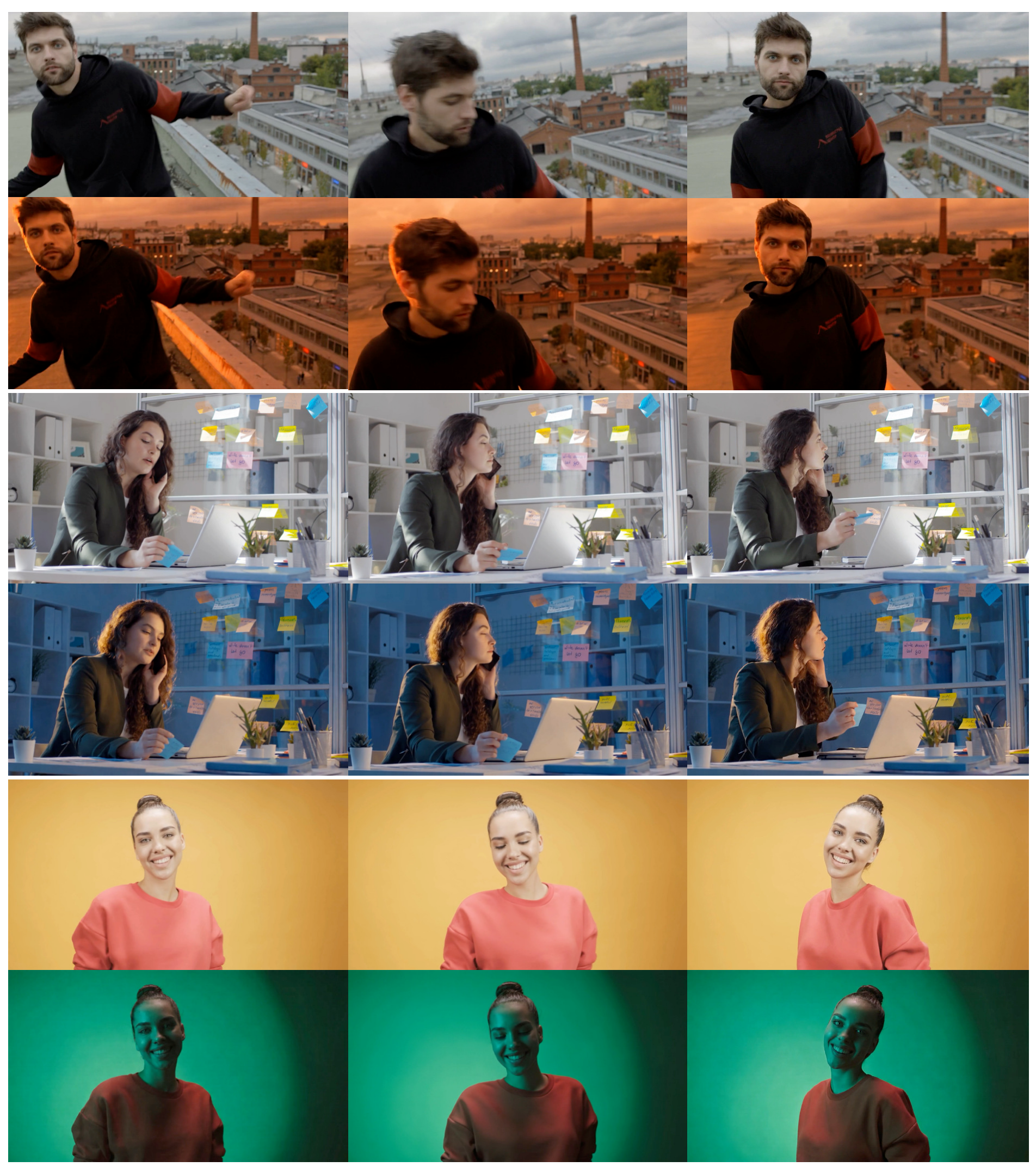}
    \caption{
    Relighting. For each pair, the first row shows the source video and the second row shows the output generated by ID-V2V. The edited keyframe modifies only the illumination while preserving the original scene content. ID-V2V consistently propagates the target lighting throughout the video, relighting not only the human subject but also the surrounding objects and background.
    }
    \label{fig:relighting_generation}
\end{figure*}